\documentclass[11pt]{article}

\usepackage[preprint]{acl}

\usepackage{times}
\usepackage{latexsym}

\usepackage[T1]{fontenc}

\usepackage[utf8]{inputenc}

\usepackage{microtype}

\usepackage{inconsolata}

\usepackage{graphicx}

\usepackage{amsmath}
\usepackage{amssymb}
\usepackage{mathtools}
\usepackage{amsthm}
\usepackage{multicol}
\usepackage{multirow}
\usepackage{algorithm}
\usepackage{algpseudocode}
\usepackage{booktabs}
\usepackage{placeins}
\title{AdaPLD: Adaptive Retrieval and Reuse for Efficient Model-Free Speculative Decoding}

\author{
Runheng Liu$^{1,*}$,
Jincheng Xie$^{2,*}$,
Wen Hu$^{3}$,
Xingchen Xiao$^{1}$,
Heyan Huang$^{1,\dagger}$ \\
$^{1}$School of Computer Science and Technology, Beijing Institute of Technology \\
$^{2}$Department of Mathematical Sciences, Tsinghua University \\
$^{3}$JDT AI Infra \\
\texttt{\{rhliu,xcxiao,hhy63\}@bit.edu.cn} \\
\texttt{xiejc22@mails.tsinghua.edu.cn} \\
\texttt{huwen.31@jd.com} \\
$^*$Equal contribution. \quad
$^\dagger$Corresponding author.
}

\begin{document}
\maketitle

\begin{abstract}
Speculative decoding accelerates generation by verifying multiple drafted tokens in a single target-model forward pass, reducing sequential decoding iterations. Model-free variants avoid auxiliary draft models by reusing text and model states already available during generation, but their speedup depends on the reliability of the constructed drafts. We identify two limitations of existing reuse-based methods: lexically anchored retrieval has limited recall under surface-form variation, and deterministic span copying can be brittle when the retrieved context does not uniquely determine the continuation. We propose \emph{AdaPLD}, a training-free method that adaptively improves both retrieval and draft construction. AdaPLD preserves high-precision lexical reuse while using semantic similarity to recover additional reuse opportunities when lexical matching fails. It further constructs branched reuse hypotheses to account for continuation uncertainty, rather than relying on a single copied span. Across diverse benchmarks, AdaPLD reduces target-model forward passes and achieves up to $3.10\times$ decoding speedup.

\end{abstract}

\section{Introduction}

Large language models (LLMs) are increasingly used for open-ended generation and instruction following, but their deployment is often limited by decoding latency. During auto-regressive decoding, each output token requires a separate target-model forward pass conditioned on the previously generated sequence, creating a strictly sequential bottleneck. Speculative decoding (SD) mitigates this bottleneck by verifying a proposed draft continuation with the target model and accepting multiple tokens within a single decoding iteration when the draft agrees with the target model \cite{10.5555/3618408.3619203,chen2023acceleratinglargelanguagemodel}. Model-free SD is a particularly practical variant because it avoids the need for an auxiliary draft model, shifting the central challenge to constructing useful drafts from information already available during generation.

Avoiding an auxiliary draft model makes draft construction the central determinant of model-free SD efficiency. Since there is no learned proposer to predict future tokens, reuse-based methods must first locate relevant context positions and then transform their continuations into drafts that the target model is likely to accept. Existing model-free SD methods can therefore be viewed as making two consecutive decisions: a retrieval decision that identifies reusable positions from the available context, and a reuse decision that constructs draft continuations from the retrieved positions \cite{saxena2023prompt,somasundaram-etal-2025-pld}. Failures in either decision lead to short accepted prefixes and limit the benefit of speculation.

The first bottleneck lies in retrieval. Most prior methods identify reuse positions through token-level matching, typically using $n$-gram heuristics. Exact matching is precise when surface overlap exists, but it has limited recall when useful reuse positions are expressed with different surface forms. PLD+ \cite{somasundaram-etal-2025-pld} improves candidate selection by reranking lexically retrieved positions with representation similarity. However, this reranking operates only after lexical candidates have been found, so it does not address cases where lexical matching fails to expose any candidate positions. This candidate-reachability constraint leads to a no-hit failure mode. When the anchor token does not appear in the available context, retrieval returns an empty candidate set, and reuse-based draft construction cannot be applied, even if semantically related context is available under different surface forms.

The second limitation lies in reuse construction. Even after a plausible reuse position is found, existing methods typically construct drafts by deterministically copying a single contiguous span from that position \cite{saxena2023prompt,hu-etal-2025-sam,somasundaram-etal-2025-pld}. This design treats the continuation following the retrieved anchor as the only reuse hypothesis. However, a locally matched anchor may support multiple plausible next-token continuations under the current context, while span copying exposes only the historical continuation observed in memory. This limits reuse construction to a single path and leaves alternative continuations unexplored.

We propose \emph{AdaPLD}, a training-free speculative decoding method that improves both stages of context reuse. For retrieval, AdaPLD keeps lexical matching as the default path and invokes semantic retrieval only when lexical matching finds no candidates, thereby expanding candidate reachability without sacrificing the precision of exact matches. For reuse construction, AdaPLD treats a selected anchor as the start of a small set of reuse hypotheses rather than a single copied span: it branches over plausible next tokens and then extends each branch with a short successor reuse step when possible.

Empirically, AdaPLD achieves consistent decoding speedups across input-guided generation, code editing, and reasoning benchmarks, with a maximum speedup of $3.10\times$. Our contributions are threefold: (1) we identify lexical no-hit retrieval as a key bottleneck in reuse-based model-free SD; (2) we propose an adaptive retrieval-and-reuse method that combines strict-first semantic retrieval with branching-based draft construction; and (3) we validate the efficiency gains of AdaPLD across diverse generation settings.

\section{Related Work}

\paragraph{Speculative decoding.}
Speculative decoding accelerates auto-regressive generation by separating draft proposal from target-model verification \cite{leviathan2023fastinferencetransformersspeculative,chen2023acceleratinglargelanguagemodel}. A draft mechanism proposes candidate future tokens, and the target model verifies these tokens in parallel, allowing multiple tokens to be accepted within a single decoding iteration when verification succeeds. A common formulation uses an auxiliary draft model to generate proposals, while model-free variants avoid an additional model and construct drafts from information already available during decoding. This work focuses on the model-free setting and improves how reusable context is identified and exploited for draft construction.

\paragraph{Model-free draft construction via context reuse.}
Model-free speculative decoding methods construct draft tokens without training a separate proposer. A common strategy is context reuse, where draft tokens are copied or derived from existing text in the input, the decoding history, or external retrieval sources \cite{saxena2023prompt,fu2024break,oliaro2025suffixdecoding,he-etal-2024-rest,ma-etal-2025-cacheback,luo-etal-2025-turning}. In these methods, draft construction is typically organized around two steps: identifying candidate reuse positions and using the tokens following those positions as draft continuations. This design removes the cost of an auxiliary draft model, but makes decoding efficiency dependent on whether relevant reuse positions can be found and whether their continuations provide useful drafts.

\paragraph{Retrieval signals for context reuse.}
Most context-reuse methods identify candidate positions through token-level matching, such as $n$-gram, suffix, or exact-token matching \cite{fu2024break,oliaro2025suffixdecoding,he-etal-2024-rest,gritta-etal-2025-dresd,ma-etal-2025-cacheback,song-etal-2025-accelerated,hu-etal-2025-sam,quan-etal-2025-rasd,10.1145/3637528.3671614}. Lexical matching is efficient and precise when exact overlaps are present, but it restricts retrieval to positions exposed by surface-form overlap. Recent work such as PLD+ \cite{somasundaram-etal-2025-pld} incorporates representation similarity to rerank lexically retrieved candidates, improving the selection of reuse anchors. However, such semantic signals are applied after lexical candidate generation, so they do not resolve cases where lexical matching returns no candidates. AdaPLD instead uses semantic retrieval as a fallback path when lexical retrieval fails, expanding candidate reachability while retaining exact matching as the default retrieval mechanism.

\section{Method}

\subsection{Problem Setup}

We consider auto-regressive generation with a target LLM. Given an input prompt $x$ and previously generated tokens $y_{<i}=(y_1,\ldots,y_{i-1})$, the target model defines the next-token distribution as
\begin{equation}
p(y_i \mid x, y_{<i}).
\end{equation}

This work focuses on model-free speculative decoding, where draft hypotheses are constructed without an auxiliary draft model. At decoding step $i$, we define the retrieval memory as the observed context
\begin{equation}
M_i = \mathrm{concat}(x, y_{<i}),
\end{equation}
which contains both the input prompt and the generated history. We use $M_{i,j}$ to denote the token at position $j$ in $M_i$, and $M_{i,\le j}$ to denote the prefix ending at position $j$.

Reuse-based draft construction retrieves positions from $M_i$ and uses the retrieved positions to form speculative hypotheses for future tokens. This unified memory allows reuse to originate from either the prompt or the generated history: input-guided tasks often reuse content from $x$, whereas reasoning-oriented generation may reuse patterns that emerge in $y_{<i}$.

\subsection{Overview of AdaPLD}

AdaPLD constructs speculative drafts by retrieving reusable positions from a unified memory and then expanding the selected reuse anchor into a small draft tree for verification. At each decoding step, AdaPLD first retrieves candidate anchor positions from the prompt and generation history, prioritizing exact lexical matches and falling back to semantic similarity only when lexical retrieval fails. The retrieved candidates are then reranked by hidden-state similarity to select an anchor that is compatible with the current decoding context.

Given the selected anchor, AdaPLD constructs draft hypotheses through three reuse operations. If the anchor is obtained by exact lexical matching, AdaPLD copies a main continuation from memory. To account for uncertainty in the immediate continuation, it further introduces branch tokens from the anchor-conditioned next-token distribution. Finally, successor drafting extends branch hypotheses by one additional reused token when possible. The resulting draft tree is verified by the target model with tree attention.

\subsection{Adaptive Retrieval with Semantic Fallback}\label{sec:retrieval}

At each decoding step, AdaPLD first retrieves candidate anchor positions from the memory $M_i$. It uses a lexical-first retrieval strategy: exact token matching is used when lexical anchors are available, and semantic fallback is activated only when no lexical anchor is found.

\paragraph{Lexical anchoring via exact token match.}

Let the query anchor at decoding step $i$ be the most recently generated token $a_i = y_{i-1}$. 
AdaPLD first retrieves earlier memory positions whose token identity matches the query anchor:
\begin{equation}
\mathcal{C}_{\text{lex}} =
\{ j \mid 1 \le j < |M_i|,\; M_{i,j} = a_i \}.
\end{equation}
This lexical anchoring step follows prior reuse-based speculative decoding methods \cite{somasundaram-etal-2025-pld} and serves as the default retrieval path because exact token matches provide high-precision anchor candidates.

\paragraph{Semantic fallback retrieval via token embedding similarity.}

If $\mathcal{C}_{\text{lex}}$ is empty, AdaPLD activates semantic fallback retrieval, using token embedding similarity as a lightweight signal to recover positions whose token representations are close to the query anchor. It compares the embedding of the query anchor $a_i$ with token embeddings at earlier memory positions and retrieves positions whose cosine similarity exceeds a threshold:
\begin{equation}
\mathcal{C}_{\text{sem}} =
\{ j \mid 1 \le j < |M_i|,\;
\cos(\mathbf{e}_{M_{i,j}}, \mathbf{e}_{a_i}) \ge \tau \}.
\end{equation}
where $\mathbf{e}_{M_{i,j}}$ and $\mathbf{e}_{a_i}$ denote token embeddings from the target model, and $\tau$ is the semantic similarity threshold. The final retrieval result is determined by a lexical--semantic backoff rule:
\begin{equation}
\mathcal{C}_i =
\begin{cases}
\mathcal{C}_{\text{lex}}, & \text{if } \mathcal{C}_{\text{lex}} \neq \emptyset, \\
\mathcal{C}_{\text{sem}}, & \text{otherwise}.
\end{cases}
\end{equation}

\subsection{Candidate Selection via Hidden-State Reranking}\label{sec:rerank}

Retrieval may return multiple candidate anchor positions in $\mathcal{C}_i$. 
AdaPLD therefore applies a selection step to choose one anchor position for downstream draft construction. 
Following \citet{somasundaram-etal-2025-pld}, AdaPLD performs this selection by hidden-state reranking. 

For each position $t$ in $M_i$, let $\mathbf{h}_{i,t}^{(\ell)}$ denote the hidden representation at layer $\ell$ after processing the prefix $M_{i,\le t}$. 
The current query anchor is the last token in the observed context, i.e., $a_i=M_{i,|M_i|}=y_{i-1}$. 
Its preceding context is therefore represented by $\mathbf{h}_{i,|M_i|-1}^{(\ell)}$. 
For a candidate anchor position $j \in \mathcal{C}_i$, the corresponding preceding context is represented by $\mathbf{h}_{i,j-1}^{(\ell)}$.

AdaPLD assigns each candidate a compatibility score by comparing these two prefix representations:
\begin{equation}
s(j) =
\cos\left(
\mathbf{h}_{i,j-1}^{(\ell)},
\mathbf{h}_{i,|M_i|-1}^{(\ell)}
\right).
\end{equation}
The selected anchor is the highest-scoring candidate:
\begin{equation}
j^\star = \arg\max_{j \in \mathcal{C}_i} s(j).
\end{equation}

\subsection{Adaptive Reuse Construction}

Given the selected anchor position $j^\star$, AdaPLD constructs a small draft tree from the memory $M_i$. The construction consists of three components: a main copy path when lexical retrieval succeeds, branch paths for alternative immediate continuations, and successor extensions for short reuse beyond branch tokens.

\paragraph{Main reuse from a lexical anchor.}

AdaPLD constructs a main copy path only when $\mathcal{C}_{\text{lex}} \neq \emptyset$. 
In this case, the selected anchor $j^\star$ is obtained from exact lexical retrieval, so AdaPLD reuses the following memory tokens as a direct continuation.
Let
\begin{equation}
\ell_i = \min(\ell_m, |M_i|-j^\star),
\end{equation}
where $\ell_m$ is the maximum copy length allowed by the speculation budget. 
The main reuse draft is
\begin{equation}
\hat{y}^{\mathrm{main}}_{i:i+\ell_i-1}
=
M_{i,j^\star+1:j^\star+\ell_i}.
\end{equation}
When the anchor is obtained via semantic fallback retrieval, no well-defined memory continuation is available, and the direct span copying is omitted.

\paragraph{Branching for local continuation uncertainty.}

To avoid relying only on the copied continuation, AdaPLD also constructs branch paths from the next-token distribution conditioned on the selected anchor prefix. 
Specifically, it obtains the top-$K$ tokens from
\begin{equation}
\widetilde{\mathcal{B}}_i
=
\mathrm{TopK}\big(p(\cdot \mid M_{i,\le j^\star}), K\big).
\end{equation}
If a main copy path exists, AdaPLD removes its first token from the branch set to avoid duplicating the same immediate continuation:
\begin{equation}
\mathcal{B}_i =
\begin{cases}
\widetilde{\mathcal{B}}_i \setminus \{M_{i,j^\star+1}\}, 
& \text{if } \mathcal{C}_{\text{lex}} \neq \emptyset,\\
\widetilde{\mathcal{B}}_i, 
& \text{otherwise}.
\end{cases}
\end{equation}
Each branch token $b\in\mathcal{B}_i$ defines a one-token draft path with
\begin{equation}
\hat{y}^{(b)}_i = b.
\end{equation}
These branch tokens are proposed from the anchor-conditioned distribution and are later verified under the current decoding context.

\paragraph{Successor drafting for extended reuse.}

A branch token only proposes the immediate next token. 
To extend a branch by one additional reused token, AdaPLD performs successor drafting for each $b\in\mathcal{B}_i$. 
It treats $b$ as a secondary query token and retrieves successor candidates from memory positions that have an observed following token. 
Using the same lexical-first backoff rule as in \S~\ref{sec:retrieval}, with $b$ replacing the original query anchor, AdaPLD obtains a candidate set $\mathcal{C}^{(b)}_i$.

The successor candidate is then selected by hidden-state reranking. 
For a candidate position $t\in\mathcal{C}^{(b)}_i$, AdaPLD compares the prefix representation before $t$ with the representation of the selected anchor prefix:
\begin{equation}
s_b(t)
=
\cos\left(
\mathbf{h}^{(\ell)}_{i,t-1},
\mathbf{h}^{(\ell)}_{i,j^\star}
\right).
\end{equation}
The successor position is
\begin{equation}
j_b^\star = \arg\max_{t\in\mathcal{C}^{(b)}_i} s_b(t).
\end{equation}
AdaPLD then extends the branch with the token following the selected successor position:
\begin{equation}
\hat{y}^{(b)}_{i+1} = M_{i,j_b^\star+1}.
\end{equation}
Successor drafting is applied once per branch, extending each branch path by at most one additional reused token.

\subsection{Verification}

All draft hypotheses constructed through main reuse, branching, and successor drafting are organized into a draft tree and verified by the target model with tree attention \cite{cai2024medusa}. 
Tree attention allows multiple draft paths to be evaluated in parallel within a single forward pass, amortizing the cost of verification.

AdaPLD follows the standard speculative decoding verification rule described in Appendix~\ref{sec:background}. 
Draft tokens are accepted only when they pass verification; otherwise, decoding falls back to the target model according to the standard rejection-sampling correction. 
Thus, AdaPLD changes only the proposal mechanism, while accepted and replacement tokens remain determined by the target model.

This verification procedure preserves the losslessness guarantee of speculative decoding while allowing AdaPLD to improve efficiency through adaptive retrieval and reuse-based draft construction.

\section{Experiments}

\subsection{Settings}

\paragraph{Datasets.}

We evaluate AdaPLD on three groups of benchmarks. First, for input-guided generation, we follow PLD+ \cite{somasundaram-etal-2025-pld} and use the summarization task \cite{nallapati-etal-2016-abstractive} and the multi-turn conversation task \cite{zheng2023judging} from Spec-Bench \cite{xia-etal-2024-unlocking}. We also include the retrieval-augmented generation (RAG) task \cite{10.1162/tacl_a_00276,karpukhin-etal-2020-dense} from Spec-Bench and treat it as input-guided generation, since the generated output is grounded in the retrieved documents provided as part of the input. 
Second, for input-guided editing, we evaluate on CodeEditorBench \cite{guo2025codeeditorbenchevaluatingcodeediting}, which covers debugging, translation, polishing, and requirement switching. These tasks are suitable for reuse-based speculative decoding because the output is often constrained by and partially derived from the input code. 
Third, beyond input-guided settings, we evaluate on reasoning-oriented benchmarks, including AIME 2024, MATH-500 \cite{lightman2024lets}, and MMLU-Pro \cite{wang2024mmlupro}. 
Although these tasks do not explicitly require copying from the input, they involve structured multi-step generation, allowing us to examine whether reuse can arise from previously generated reasoning traces.


\paragraph{Models.}

We conduct experiments with Vicuna-v1.3 models \cite{zheng2023judging} at three scales: 7B, 13B, and 33B parameters.
This follows prior speculative decoding studies and allows us to evaluate whether the speedup trend is consistent across model sizes.

\paragraph{Baselines.}

We compare our method against a set of representative baselines, including standard auto-regressive decoding (AR), PLD \cite{saxena2023prompt}, PLD+ \cite{somasundaram-etal-2025-pld}, Token Recycling (TR) \cite{luo-etal-2025-turning}, SAM Decoding (SAMD) \cite{hu-etal-2025-sam}, and Logitspec \cite{liu2026logitspecacceleratingretrievalbasedspeculative}. All methods are evaluated with the same target model and without auxiliary draft models, so the comparison focuses on differences in draft construction. These methods cover both classical and recent model-free speculative decoding approaches, allowing for a comprehensive evaluation of efficiency and effectiveness.

\paragraph{Metrics.}

We report end-to-end decoding throughput measured in tokens per second, as well as overall speedup relative to standard auto-regressive decoding.

\paragraph{Implementation Details.}

All experiments are conducted using half-precision (FP16) inference on 8 NVIDIA H20 GPUs. For each baseline, we use the recommended hyperparameters from the original implementation whenever applicable. Unless otherwise stated, experiments are run with temperature $T=0$ to ensure deterministic decoding and stable reuse behavior. For hidden-state reranking, we follow PLD+ and use hidden states from the same transformer layer without tuning the layer choice (layer indices in Table~\ref{tab:params}).

\subsection{Main Results}

\begin{table}[t]
    \centering
\resizebox{0.5\textwidth}{!}{
    \begin{tabular}{ll|ccc|cc}
    \toprule
         &  \bf Method& \bf MT& \bf Sum&\bf RAG& \bf \#tokens/s&\bf Avg.\\
         \midrule
         \multirow{7}{*}{\rotatebox[origin=c]{90}{Vicuna-7B}}& AR& 1.00& 1.00& 1.00& 59.55&1.00\\
         & TR& \bf 2.05& 1.81& 1.61& 109.11&1.83\\
         & SAMD& 1.70& 2.64& 1.70& 120.39&2.02\\
         & PLD& 1.39& 2.18& 1.53& 101.33&1.70\\
         & PLD+& 1.63& 2.77& 1.71& 121.72&2.04\\
         & Logitspec& 1.88& 2.59& 1.70& 123.17&2.07\\
         & Ours& 1.83& \bf 3.02& \bf 1.93& \bf 135.08&\bf 2.27\\
 
         \midrule
         
         \multirow{7}{*}{\rotatebox[origin=c]{90}{Vicuna-13B}}& AR& 1.00& 1.00& 1.00& 47.87&1.00\\
         & TR& 1.48& 1.32& 1.26& 65.06&1.36\\
         & SAMD& 1.60& 2.17& 1.62& 86.30&1.80\\
         & PLD& 1.34& 1.90& 1.46& 75.09&1.57\\
         & PLD+& 1.50& 2.22& 1.60& 85.09&1.78\\
         & Logitspec& 1.65& 2.03& 1.60& 84.40&1.76\\
         & Ours& \bf 1.70& \bf 2.45& \bf 1.79& \bf 94.88&\bf 1.98\\
         \midrule
         \multirow{7}{*}{\rotatebox[origin=c]{90}{Vicuna-33B}}& AR& 1.00& 1.00& 1.00& 27.43&1.00\\
         & TR& 1.25& 1.22& 1.15& 33.18&1.21\\
         & SAMD& 1.53& 1.91& 1.41& 44.34&1.62\\
         & PLD& 1.28& 1.66& 1.19& 37.85&1.38\\
         & PLD+& 1.48& 1.95& 1.38& 43.98&1.60\\
         & Logitspec& 1.30& 1.65& 1.22&  38.10&1.39\\
         & Ours& \bf 1.62& \bf 2.10& \bf 1.50& \bf 47.70& \bf 1.74\\
         \bottomrule
 
     \end{tabular}
     }
     \caption{Normalized speedup over auto-regressive decoding (AR) on input-guided generation tasks from Spec-Bench.}
    \label{tab:ig_generation_vicuna}
\end{table}
\begin{table}[th]
    \centering

\resizebox{0.5\textwidth}{!}{
    \begin{tabular}{ll|cccc|cc}
    \toprule
         &  \bf Method& \bf Debug&\bf Polish&\bf Translate&\bf Switch& \bf \#tokens/s&\bf Avg.\\
         \midrule
         \multirow{7}{*}{\rotatebox[origin=c]{90}{Vicuna-7B}}& AR& 1.00& 1.00&1.00& 1.00& 64.17&1.00\\
         & TR&  2.03&  2.01&  \bf 2.06&  2.05& 131.03&2.04\\
 & SAMD&  1.94&  1.92&  1.53&  2.05& 120.23&1.87\\
         & PLD&  1.60&  1.63&  1.31&  1.70& 100.23&1.56\\
         & PLD+&  1.94&  1.93&  1.55&  2.04& 119.96&1.87\\
 & Logitspec&  \bf 2.17&  \bf 2.22&  1.70&  \bf 2.22& \bf 133.49&\bf 2.08\\
         & Ours&  2.07&  2.04&  1.70&  2.18& 128.30&2.00\\
 
         \midrule
         
         \multirow{7}{*}{\rotatebox[origin=c]{90}{Vicuna-13B}}& AR& 1.00& 1.00& 1.00& 1.00& 51.03&1.00\\
         & TR& 1.53& 1.53& 1.52& 1.57& 78.59&1.54\\
 & SAMD& 2.00& 2.18& 1.51& 4.38& 128.36&2.52\\
         & PLD& 1.57& 1.64& 1.24& 2.68& 90.82&1.78\\
         & PLD+& 1.99& 2.04& 1.48& 4.15& 123.07&2.41\\
 & Logitspec& 1.89& 2.12& 1.52& 3.31& 112.62&2.21\\
         & Ours& \bf 2.28& \bf 2.34& \bf 1.70& \bf 4.51& \bf 138.10&\bf 2.71\\
         \midrule
         \multirow{7}{*}{\rotatebox[origin=c]{90}{Vicuna-33B}}& AR& 1.00& 1.00&1.00& 1.00& 29.97&1.00\\
         & TR& 1.33& 1.32& 1.39& 1.37& 40.57&1.35\\
 & SAMD& 2.47& 2.82& 1.63& 4.47& 85.32&2.85\\
         & PLD& 1.88& 2.08& 1.37& 3.12& 63.30&2.11\\
         & PLD+& 2.44& 2.79& 1.68& 4.73& 87.05&2.90\\
 & Logitspec& 1.97& 2.17& 1.43& 3.01& 64.27&2.14\\
         & Ours& \bf 2.77& \bf 2.99& \bf 1.84& \bf 4.83& \bf 93.04&\bf 3.10\\
         \bottomrule
 
     \end{tabular}
     }
     \caption{Normalized speedup over auto-regressive decoding (AR) on input-guided editing tasks from CodeEditorBench.}
    \label{tab:ig_editing_code_vicuna}
\end{table}
\begin{table}[]
    \centering

    \setlength{\tabcolsep}{1mm}
\resizebox{0.5\textwidth}{!}{
    \begin{tabular}{ll|ccc}
    \toprule
         &  \bf Method&\bf AIME 2024&\bf MATH-500&\bf MMLU-Pro \\
         \midrule
         \multirow{7}{*}{\rotatebox[origin=c]{90}{Vicuna-7B}}& AR& 1.00  &1.00 &1.00   \\
         & TR&  2.22&  2.36&  2.18\\
 & SAMD&  2.78&  2.32&  2.14\\
         & PLD&  2.01&  1.74&  1.68\\
         & PLD+&  2.61&  2.42&  2.12\\
 & Logitspec&  \bf 3.15&  \bf 2.68&  1.94\\
         & Ours&  2.98&  2.55&  \bf 2.44\\
 
         \midrule
         
         \multirow{7}{*}{\rotatebox[origin=c]{90}{Vicuna-13B}}& AR& 1.00  & 1.00 &1.00   \\
         & TR& 1.64 & 1.65 &1.59  \\
 & SAMD& 2.34 & 1.95 &1.79  \\
         & PLD& 1.73 & 1.64 &1.47  \\
         & PLD+& 2.36& 1.98&1.81\\
 & Logitspec& 2.33 & 2.12 &1.54  \\
         & Ours& \bf 2.56& \bf 2.26&\bf 2.08\\
         \midrule
         \multirow{7}{*}{\rotatebox[origin=c]{90}{Vicuna-33B}}& AR& 1.00  &1.00 &1.00   \\
         & TR& 1.37 & 1.41 &1.29  \\
 & SAMD& 1.84 & 1.75 &1.67  \\
         & PLD& 1.50 & 1.53 &1.42  \\
         & PLD+& 1.86& 1.83& 1.64\\
 & Logitspec& 1.60 & 1.64 &1.46  \\
         & Ours& \bf 2.10& \bf 2.02& \bf 1.81\\
         \bottomrule
 
     \end{tabular}
     }
     \caption{Normalized speedup over auto-regressive decoding (AR) on reasoning benchmarks.}
    \label{tab:reasoning_vicuna}
\end{table}

\paragraph{Results on input-guided generation tasks.}

Table~\ref{tab:ig_generation_vicuna} summarizes the speedup results of AdaPLD on input-guided generation tasks from Spec-Bench. Across all evaluated model scales, AdaPLD achieves the highest average speedup among reuse-based speculative decoding methods. Specifically, AdaPLD attains an average speedup of $2.27\times$ on Vicuna-7B, $1.98\times$ on Vicuna-13B, and $1.74\times$ on Vicuna-33B. Compared to PLD+, AdaPLD yields consistent relative improvements of $8$--$11\%$ in average speedup across model sizes. 
At the task level, AdaPLD consistently outperforms prior methods on summarization and retrieval-augmented generation, where the generated output is highly grounded in the input context. For multi-turn conversation, AdaPLD remains competitive across model scales and achieves the highest speedup on larger models. This behavior is expected, as prior reuse-based speculative decoding methods such as PLD and PLD+ also exhibit relatively limited gains on multi-turn conversation. This behavior may be related to the structure of multi-turn conversation, where the initial turn is closer to free-form generation. From the second turn onward, the accumulated dialogue history serves as input, enabling reuse-based speculative decoding to take effect.

\paragraph{Results on input-guided editing tasks.}

Table~\ref{tab:ig_editing_code_vicuna} presents the results on input-guided editing tasks from CodeEditorBench. AdaPLD achieves the highest average speedup on the Vicuna-13B and Vicuna-33B models, with average speedups of $2.71\times$ and $3.10\times$, respectively. On the Vicuna-7B model, Logitspec achieves the highest average speedup, while AdaPLD remains competitive.
At the task level, AdaPLD consistently outperforms prior methods on debugging, polishing, and requirement switching across the 13B and 33B models. For translation, where the output is less directly constrained by the input structure, all methods exhibit lower speedups, though AdaPLD remains among the top-performing approaches. Overall, these results indicate that AdaPLD is particularly effective for editing tasks with strong input grounding and high reuse potential.

\paragraph{Results on reasoning tasks.}

Table~\ref{tab:reasoning_vicuna} reports the speedup results on reasoning-oriented benchmarks. Although these tasks are not explicitly input-guided, AdaPLD achieves competitive speedups across model scales, and often outperforms prior methods on larger models. In particular, AdaPLD attains the highest speedup on all three reasoning benchmarks for the Vicuna-13B and Vicuna-33B models. On the Vicuna-7B model, AdaPLD
remains competitive but is slightly outperformed by Logitspec on AIME~2024 and MATH-500, while achieving the best result on MMLU-Pro. These results suggest that reuse-based speculative decoding can still be effective in reasoning scenarios, where reuse may arise from structured and incremental reasoning processes rather than direct input overlap.

\subsection{Ablation Study}

\begin{table*}[t]
    \centering
\resizebox{\textwidth}{!}{
    \begin{tabular}{l|ccc|cccc|c}
    \toprule
           \bf Variant&  \bf MT &\bf SUM &\bf RAG&\bf Debug& \bf Polish & \bf Translate&\bf Switch &\bf Avg.\\
         \midrule
          Full AdaPLD &  1.70&  2.45&  1.79&   2.28& 2.34& 1.70&4.51 &2.40\\
    w/o branch and successor drafting&  1.53&  2.25&  1.64&   2.03& 2.07& 1.51&4.22 &2.18\\
    w/o successor drafting&  1.60&  2.41&  1.77&   2.26& 2.34& 1.69&4.51 &2.37\\
    w/o embedding retrieval (lexical-only) &  1.62&  2.37&  1.74&   2.20& 2.23& 1.63&3.91 &2.24\\
         \bottomrule
 
     \end{tabular}
     }
         \caption{Component ablations of AdaPLD on representative input-guided tasks.
We report normalized speedup over auto-regressive decoding (AR) on Vicuna-13B.}
    \label{tab:ablation_component}
\end{table*}

\subsubsection{Component Ablation}

Table~\ref{tab:ablation_component} reports component-level ablations of AdaPLD on representative input-guided tasks. 
All ablated variants reduce the average speedup compared with the full method, indicating that the proposed components contribute to AdaPLD's overall efficiency.

\paragraph{Branch and successor drafting.}

Removing both branch and successor drafting produces the largest average drop, reducing speedup from $2.40\times$ to $2.18\times$. 
This result shows that the selected anchor provides useful reuse signals beyond the single span directly copied from memory. 
Branch drafting exploits these signals by expanding the set of plausible continuations supported by the retrieved anchor. 
Successor drafting further increases the reuse depth of branched hypotheses, allowing AdaPLD to benefit from reusable context beyond the first branch token.

\paragraph{Successor drafting.}
Removing successor drafting alone reduces the average speedup from $2.40\times$ to $2.37\times$. 
This confirms that successor drafting provides an additional extension benefit on top of branch drafting. 
By extending branch hypotheses with reused successor tokens, this component helps AdaPLD convert local continuation alternatives into longer speculative drafts.

\paragraph{Token-embedding-based fallback retrieval.}

Disabling token-embedding-based fallback retrieval and relying only on lexical matching reduces the average speedup from $2.40\times$ to $2.24\times$. 
The largest drop occurs on requirement switching, where speedup decreases from $4.51\times$ to $3.91\times$. 
This result highlights the value of semantic fallback in expanding retrieval coverage beyond exact token matches. 
By recovering candidate anchors when lexical matching is sparse, fallback retrieval provides additional reuse opportunities for downstream draft construction.

\subsubsection{Hyperparameter Sensitivity}

\paragraph{Impact of branch width.}

\begin{figure}[th]
    \centering
    \includegraphics[width=\linewidth]{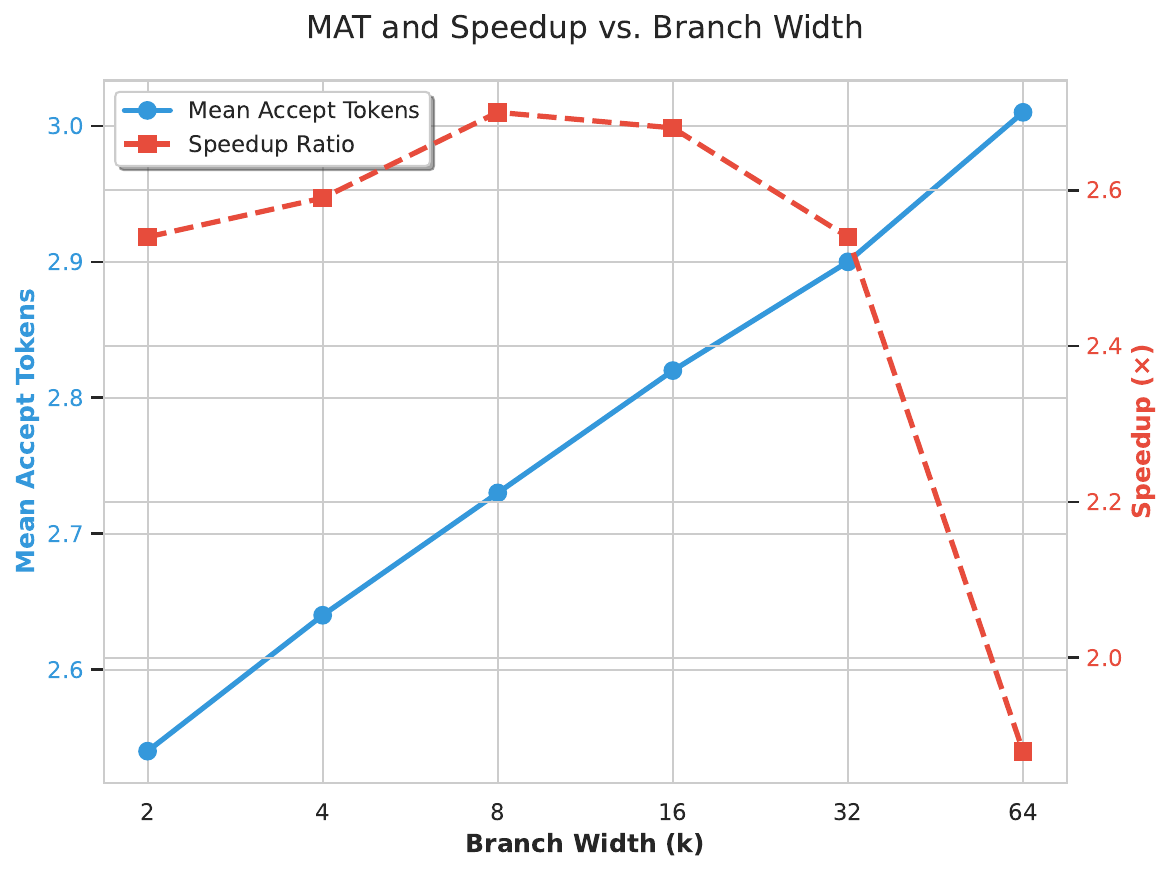}
    \caption{Effect of branch width $k$ on mean accepted tokens (left axis) and normalized speedup (right axis) for CodeEditorBench on Vicuna-13B.}
    \label{fig:abl_branch_width}
\end{figure}

We study the effect of branch width $k$ on decoding efficiency by varying $k \in \{2, 4, 8, 16, 32, 64\}$ on CodeEditorBench with Vicuna-13B. As shown in Figure~\ref{fig:abl_branch_width}, increasing the branch width leads to a increase in the average number of accepted tokens, indicating that
considering more branch candidates improves the chance of constructing reusable
drafts.
In contrast, the overall speedup exhibits a non-monotonic trend. Speedup improves as $k$ increases from 2 to 8, reaches its best performance around $k=8\sim16$, and degrades for larger values, with a clear drop observed at $k=64$. 
This behavior reflects a trade-off between reuse and verification cost: while a larger branch width increases the likelihood of identifying reusable continuations, it also incurs higher verification overhead, which eventually outweighs the benefit of additional reuse.
We use $k=8$ as the default setting in all experiments, as it lies within this plateau region while incurring lower verification overhead.

\paragraph{Impact of maximum copy length.}

\begin{figure}[th]
    \centering
    \includegraphics[width=\linewidth]{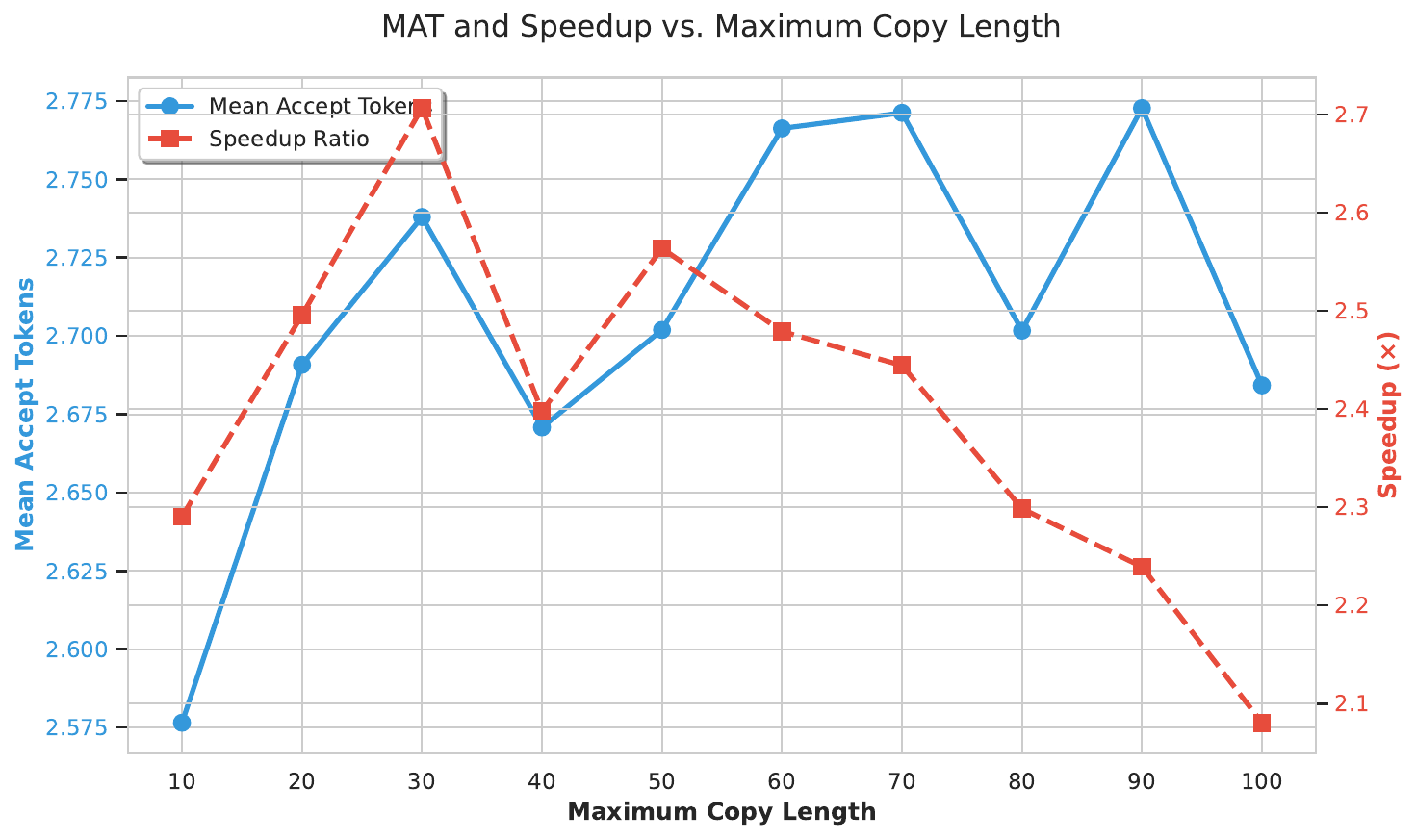}
    \caption{Effect of the maximum copy length on mean accepted tokens (left axis) and normalized speedup (right axis) for CodeEditorBench on Vicuna-13B.}
    \label{fig:abl_max_copy_length}
\end{figure}

We further study the effect of the maximum draft length by varying it from 10 to 100 on CodeEditorBench with Vicuna-13B. As shown in Figure~\ref{fig:abl_max_copy_length}, increasing the maximum draft length from 10 to 30 leads to higher numbers of accepted tokens, while further increasing the draft length results in fluctuations without a consistent upward trend. Meanwhile, the overall speedup reaches its highest observed value at a draft length of 30 and decreases overall for larger values, despite minor fluctuations. Longer drafts therefore do not consistently improve reuse, while incurring higher verification costs. We therefore set the maximum draft length to $d=30$ in all other experiments.

\paragraph{Sensitivity to semantic retrieval threshold.}

\begin{figure}[th]
    \centering
    \includegraphics[width=\linewidth]{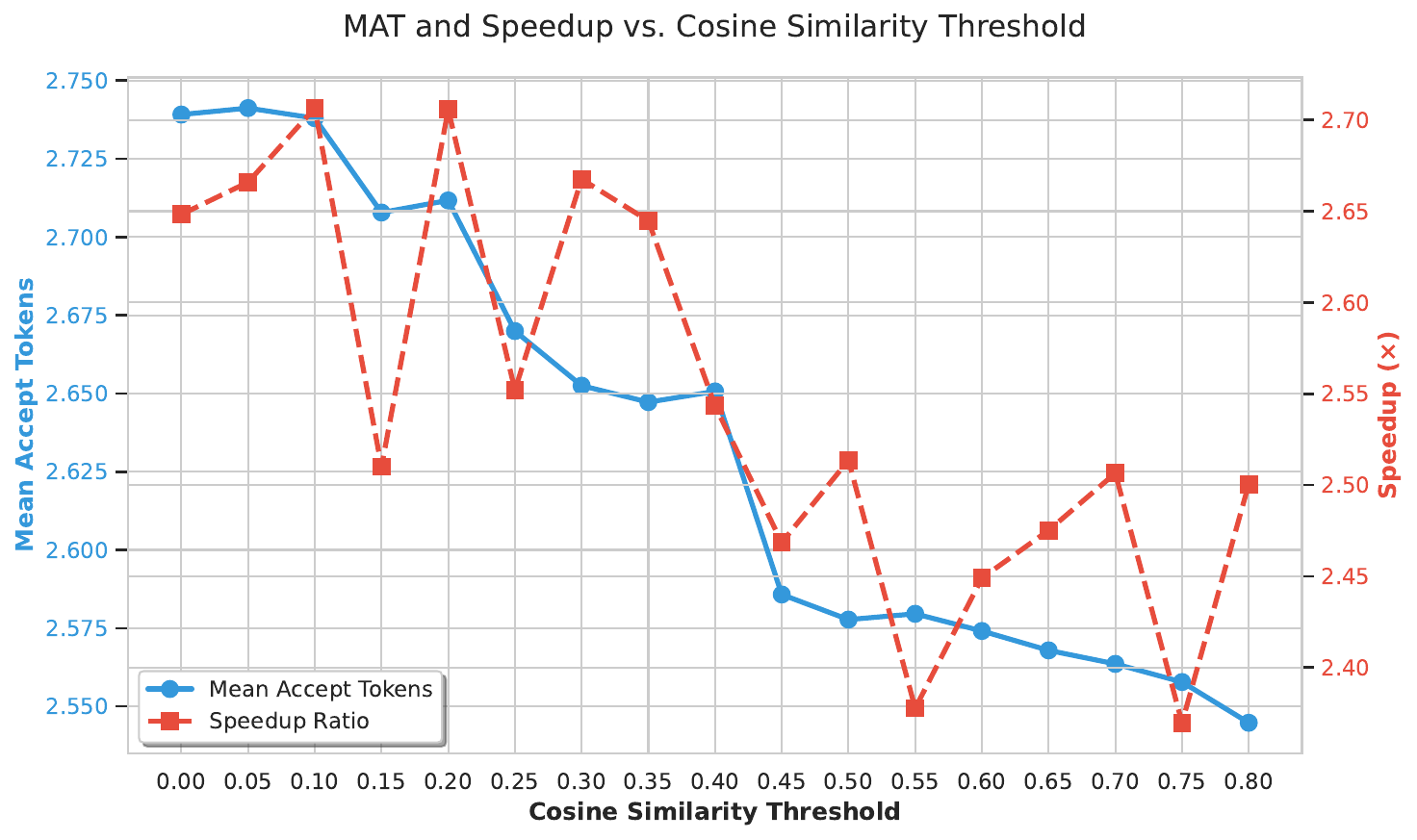}
    \caption{Effect of embedding similarity threshold on mean accepted tokens (left axis) and normalized speedup (right axis) for CodeEditorBench on Vicuna-13B.}
    \label{fig:abl_emb_sim_threshold}
\end{figure}

We analyze the sensitivity of AdaPLD to the similarity threshold used in semantic retrieval by varying it from 0.0 to 0.8 on CodeEditorBench with Vicuna-13B. As shown in Figure~\ref{fig:abl_emb_sim_threshold}, increasing the threshold generally leads to a decrease in the average number of accepted tokens, while the overall speedup exhibits non-monotonic variations across different threshold values.
Lower to moderate threshold values achieve comparable speedup, whereas higher thresholds are more often associated with reduced performance. Based on these observations, we use $\tau = 0.1$ in the remaining experiments.

\subsection{Analysis of Reuse Behavior}

\begin{figure}[th]
    \centering
    \includegraphics[width=\linewidth]{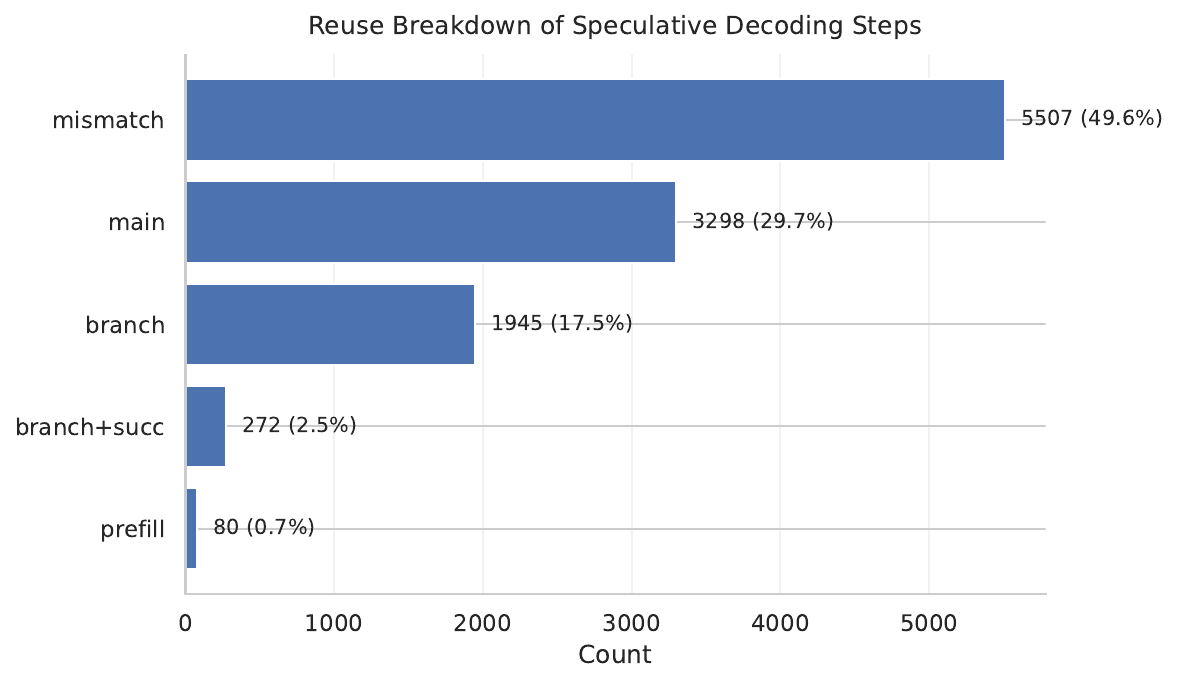}
    \caption{Reuse breakdown of speculative decoding steps for AdaPLD on CodeEditorBench (Vicuna-13B).
}
    \label{fig:ana_reuse_breakdown}
\end{figure}

We analyze reuse behavior by assigning each speculative decoding iteration to the draft path that contributes accepted tokens. 
We use three successful reuse categories: \textit{main} denotes accepted reuse from the direct span copied after the selected anchor; \textit{branch} denotes accepted reuse from an alternative branch continuation; and \textit{branch+succ} denotes accepted branch reuse that is further extended by successor drafting. 
For completeness, we also report \textit{mismatch}, where no constructed draft path contributes accepted tokens, and \textit{prefill}, where speculative decoding is not applied.

Figure~\ref{fig:ana_reuse_breakdown} shows that successful reuse is not dominated by a single mechanism. 
The main path remains a central source of accepted drafts, confirming that direct reuse from retrieved anchors is effective when the copied continuation matches the current context. 
At the same time, branch-based outcomes form a substantial portion of successful reuse, showing that the retrieved anchor contains useful continuation information beyond its observed next span. 
The presence of \textit{branch+succ} further shows that branch continuations can sometimes be extended into longer reusable hypotheses, which supports the role of successor drafting in increasing reuse depth after branching.

\section{Conclusion}

We propose AdaPLD, a model-free speculative decoding method that improves reuse-based draft construction through adaptive retrieval and structured reuse. 
AdaPLD uses lexical matching as the default retrieval path and selectively falls back to semantic similarity when exact matches are unavailable. 
It further augments direct span copying with branch and successor drafting, enabling retrieved anchors to support richer continuation hypotheses beyond a single copied span.
Experiments across input-guided generation, editing, and reasoning-oriented benchmarks show that AdaPLD improves decoding efficiency over prior model-free approaches. 
These results highlight the importance of adaptive retrieval and reuse construction for efficient model-free speculative decoding.

\section*{Limitations}

AdaPLD focuses on model-free speculative decoding, where draft construction uses information already available from the target model and the observed context. 
This design avoids training or serving an auxiliary draft model, but also means that the attainable speedup depends on whether the prompt or generation history contains reusable anchors and continuations. For tasks where the output is weakly grounded in the available context, reuse opportunities may be naturally less frequent.

AdaPLD uses token-level embedding similarity as a lightweight fallback signal when exact lexical retrieval fails. This choice keeps retrieval efficient and compatible with the model-free setting. However, token-level similarity is a relatively simple semantic signal, and future work may explore richer retrieval criteria that capture phrase-level or structural similarity while preserving low overhead.

The method also involves a small set of draft-construction hyperparameters, such as branch width, reuse length, successor depth, and embedding-similarity threshold. In this work, we use fixed settings across the main experiments to evaluate the general effectiveness of the proposed design. Different deployment environments may benefit from further tuning these parameters to match specific latency budgets, model architectures, or workload characteristics.

Finally, our experiments evaluate AdaPLD on representative model-free speculative decoding benchmarks and report decoding throughput and speedup. Since AdaPLD follows standard speculative verification, it preserves the target model's output distribution. A broader systems study on highly optimized serving frameworks would be a useful direction for understanding its end-to-end behavior in production-scale inference.

\section*{Ethical Statement}

This work studies inference-time acceleration for large language models. 
AdaPLD does not train an additional model, collect new user data, or modify the target model parameters. 
All drafted tokens are verified by the target model through standard speculative decoding, so the method preserves the target model's output distribution.

Our experiments are conducted on public benchmarks and focus on decoding efficiency. 
Since AdaPLD only changes how draft tokens are proposed before target-model verification, it does not introduce an independent generation policy beyond the underlying target model.


\bibliography{custom}

\appendix

\appendix

\section{Benchmark and Experimental Details}

\begin{table}[h]
    \centering
    \resizebox{0.5\textwidth}{!}{
    \begin{tabular}{ccc}
    \toprule
          Dataset&Subtask&  \# Samples\\
         \midrule
   \multirow{3}{*}{Spec-Bench} &MT-bench& 80\\
  &Summarization& 80\\
  &RAG&  80\\
  \midrule
 \multirow{4}{*}{CodeEditorBench}& Debug &20\\

  &Polish&  20\\
 
  &Translate&  20\\
   &Switch&  20\\
   \midrule
  MATH-500&-&  500\\
  AIME 2024&-&  30\\
  MMLU-Pro&-&  70\\

 \bottomrule
    \end{tabular}
    }
       \caption{Statistics of Benchmarks.}
    \label{tab:statistic_of_benchmarks}
\end{table}

\FloatBarrier

\begin{table}[th]
    \centering
    \begin{tabular}{lc}
    \toprule
        Model& Layer \\
        \midrule
        Vicuna-7B& 9\\
          Vicuna-13B& 13\\
 Vicuna-33B& 11\\
 Qwen3-8B&29\\
 \bottomrule
    \end{tabular}
    \caption{Transformer layer indices used for hidden-state reranking in each model.}
    \label{tab:params}
\end{table}
\FloatBarrier

\section{Additional Related Work}

\paragraph{Efficient inference.}

Efficient LLM inference has also been studied through several directions beyond speculative decoding. KV-cache compression reduces memory usage and attention overhead by retaining or compressing only a subset of cached states \cite{zhang2023ho,li2024snapkv}. Quantization lowers serving cost by representing model weights or activations with reduced precision \cite{frantar2023optq,lin2023awq}. Knowledge distillation transfers generation ability from larger models to smaller ones to reduce inference latency \cite{hinton2015distillingknowledgeneuralnetwork}. Prompt compression shortens the input context before inference, reducing the cost of prefilling and subsequent attention computation \cite{jiang-etal-2023-llmlingua,jiang-etal-2024-longllmlingua,fei-etal-2024-extending,fei2026efficient}. These methods mainly reduce per-step computation, memory footprint, and input-processing overhead, while speculative decoding mitigates the auto-regressive decoding bottleneck by verifying multiple drafted tokens in a single target-model forward pass. AdaPLD is complementary to these directions, as it improves draft construction within model-free speculative decoding.

\section{Background}\label{sec:background}

\paragraph{Speculative decoding.}

Given an input prompt $x$ and previously generated tokens $y_{<i}$, standard auto-regressive decoding generates the next token by running the target model once and sampling from $p(\cdot \mid x,y_{<i})$. 
This sequential process can underutilize parallel computation during decoding.
Speculative decoding accelerates generation by separating draft proposal from target-model verification \citep{leviathan2023fastinferencetransformersspeculative,chen2023acceleratinglargelanguagemodel}. A draft mechanism proposes one or more candidate continuations, and the target model verifies the proposed tokens in parallel. 
For a single draft path $s_{1:k}$, let $c_t=(x,y_{<i},s_{<t})$ denote the prefix before verifying $s_t$. 
We write $p_t(\cdot)=p(\cdot\mid c_t)$ and $q_t(\cdot)=q(\cdot\mid c_t)$. 
The draft token $s_t$ is accepted with probability
\begin{equation}
\alpha_t=\min\left(1,\frac{p_t(s_t)}{q_t(s_t)}\right).
\end{equation}
Verification proceeds until the first rejection, where a correction token is sampled from the residual distribution proportional to $[p_t-q_t]_+$. 
In greedy decoding, this verification reduces to accepting draft tokens that match the target model's selected tokens.
By verifying multiple proposals in one target-model pass, speculative decoding can reduce the number of sequential decoding steps. 
The rejection-sampling correction ensures that the marginal output distribution matches direct sampling from the target model, making speculative decoding lossless with respect to the target model's generation distribution \citep{leviathan2023fastinferencetransformersspeculative,chen2023acceleratinglargelanguagemodel}.

\section{Additional Evaluation}
\label{app:additional_eval}

\subsection{Stochastic Decoding Evaluation}
\label{app:sampling_decoding}

We further evaluate AdaPLD under stochastic decoding to examine whether its efficiency gains transfer beyond the deterministic setting used in the main experiments. 
We conduct experiments on CodeEditorBench with temperature $T=1$ and report normalized speedup over standard auto-regressive decoding.

We keep the AdaPLD draft-construction hyperparameters the same as in the main experiments, including the branch width, reuse length, successor depth, and embedding-similarity threshold. 
Only the hidden-state reranking layer is model-specific, and its index is reported in Table~\ref{tab:params}; the Qwen3-8B layer is selected using the same sweep protocol as PLD+ \cite{somasundaram-etal-2025-pld}.

\begin{table}[t]
\centering
\resizebox{0.5\textwidth}{!}{
\begin{tabular}{lccccc}
\toprule
Method & Debug & Polish & Translate & Switch & Avg. \\
\midrule
TR        & 1.35 & 1.27 & 1.38 & 1.33 & 1.33 \\
SAMD      & 2.55 & 2.12 & 1.57 & 3.00 & 2.31 \\
PLD       & 1.95 & 1.93 & 1.33 & 2.43 & 1.91 \\
PLD+      & 2.56 & 2.63 & 1.65 & 3.30 & 2.54 \\
Logitspec & 1.94 & 2.14 & 1.42 & 3.12 & 2.16 \\
AdaPLD    & 2.90 & 2.37 & 1.78 & 3.30 & 2.59 \\
\bottomrule
\end{tabular}
}
\caption{Normalized speedup over auto-regressive decoding under sampling-based decoding on CodeEditorBench with Vicuna-33B at temperature $T=1$.}
\label{tab:sampling_decoding_vicuna}
\end{table}
\begin{table}[t]
\centering
\resizebox{0.5\textwidth}{!}{
\begin{tabular}{lccccc}
\toprule
Method & Debug & Polish & Translate & Switch & Avg. \\
\midrule
TR        & 2.67& 2.69& 2.65& 2.55& 2.64\\
SAMD      & 2.58& 4.54& 1.67& 3.74& 3.13\\
PLD       & 2.16& 3.37& 1.40& 2.56& 2.37\\
PLD+      & 2.66& 5.51& 1.71& 4.02& 3.48\\
Logitspec & 2.87& 4.68& 1.91& 3.82& 3.32\\
AdaPLD    & 3.04& 5.65& 1.92& 4.50& 3.78\\
\bottomrule
\end{tabular}
}
\caption{Normalized speedup over auto-regressive decoding under sampling-based decoding on CodeEditorBench with Qwen3-8B at temperature $T=1$.}
\label{tab:sampling_decoding_qwen}
\end{table}

Table~\ref{tab:sampling_decoding_vicuna} reports stochastic decoding results with Vicuna-33B. 
AdaPLD achieves the highest average speedup of $2.59\times$, slightly outperforming PLD+ at $2.54\times$. 
This indicates that AdaPLD remains effective when decoding is performed with stochastic sampling rather than greedy selection.

Table~\ref{tab:sampling_decoding_qwen} reports the corresponding results with Qwen3-8B \cite{yang2025qwen3technicalreport}. 
AdaPLD achieves the best average speedup of $3.78\times$, improving over PLD+ at $3.48\times$. 
Since the draft-construction hyperparameters are kept the same as in the main experiments, these results suggest that AdaPLD's reuse strategy transfers to stochastic decoding and remains effective across target models.

\subsection{Computational Overhead}
\label{app:overhead}

We profile the computational overhead of AdaPLD on CodeEditorBench with Vicuna-33B. 
For each component, we report its latency when triggered and its average latency per speculative decoding step, where the latter accounts for the component's trigger frequency. 
We also normalize the per-step latency by the latency of the target-model verification forward pass.

\begin{table*}[t]
\centering
\resizebox{\textwidth}{!}{
\begin{tabular}{lccc}
\toprule
Component & Triggered latency (ms) & Avg. latency / step (ms) & \% of verify forward \\
\midrule
Verify forward              & 39.195 & 39.195 & 100.00 \\
Semantic fallback retrieval & 0.107  & 0.028  & 0.07 \\
Hidden-state reranking      & 0.052  & 0.049  & 0.12 \\
Branch construction         & 0.069  & 0.066  & 0.17 \\
Successor drafting          & 0.322  & 0.069  & 0.17 \\
\bottomrule
\end{tabular}
}
\caption{Component-level latency profiling of AdaPLD on CodeEditorBench with Vicuna-33B. The average latency per step accounts for the trigger frequency of each component.}

\label{tab:overhead}
\end{table*}

Table~\ref{tab:overhead} shows that AdaPLD's additional retrieval and draft-construction components are lightweight compared with target-model verification. 
Semantic fallback retrieval, hidden-state reranking, branch construction, and successor drafting each contribute less than $0.2\%$ of the verification-forward latency on average. 
Together, these components add $0.212$ ms per speculative step, corresponding to $0.53\%$ of the verification-forward cost. This profiling indicates that AdaPLD's adaptive retrieval and reuse construction introduce only minor component-level overhead compared with target-model verification.

\section{Additional Analysis}
\label{app:additional_analysis}

\subsection{Retrieval Hit Rate}
\label{app:retrieval_hit_rate}

We further analyze the retrieval outcomes of AdaPLD on Vicuna-33B across different benchmarks. 
For each retrieval attempt, the outcome is assigned to one of three categories. 
A lexical hit occurs when exact token matching retrieves at least one candidate. 
If lexical retrieval fails, semantic fallback is applied; the attempt is counted as a semantic hit if fallback retrieves at least one candidate, and as no-hit otherwise.

Let $n_{\mathrm{lex}}$, $n_{\mathrm{sem}}$, and $n_{\mathrm{none}}$ denote the number of lexical hits, semantic hits, and no-hit cases, respectively. 
The total number of retrieval attempts is
\begin{equation}
N = n_{\mathrm{lex}} + n_{\mathrm{sem}} + n_{\mathrm{none}} .
\end{equation}
We report
\begin{equation}
\begin{aligned}
R_{\mathrm{lex}} &= \frac{n_{\mathrm{lex}}}{N}, &
R_{\mathrm{sem}} &= \frac{n_{\mathrm{sem}}}{N}, \\
R_{\mathrm{none}} &= \frac{n_{\mathrm{none}}}{N}, &
R_{\mathrm{rec}} &= 
\frac{n_{\mathrm{sem}}}
{n_{\mathrm{sem}} + n_{\mathrm{none}}}.
\end{aligned}
\end{equation}
Here $R_{\mathrm{rec}}$ measures the fraction of lexical-miss cases recovered by semantic fallback.

\begin{table}[t]
\centering
\resizebox{0.5\textwidth}{!}{
\begin{tabular}{lcccc}
\toprule
\textbf{Benchmark} & \textbf{Lex. Hit} & \textbf{Sem. Hit} & \textbf{No-hit} & \textbf{Sem. Rec.} \\
\midrule
Spec-Bench      & 61.17 & 26.16 & 12.67 & 67.37 \\
CodeEditorBench & 73.43 & 21.28 & 5.30  & 80.07 \\
AIME24          & 72.89 & 20.61 & 6.50  & 76.01 \\
MATH-500        & 73.26 & 20.55 & 6.19  & 76.84 \\
MMLU-Pro        & 64.33 & 24.41 & 11.27 & 68.41 \\
\bottomrule
\end{tabular}
}
\caption{Retrieval hit-rate analysis on Vicuna-33B. All values are percentages. Sem. Rec. denotes semantic recovery among lexical-miss cases.}
\label{tab:retrieval_hit_rate}
\end{table}

Table~\ref{tab:retrieval_hit_rate} shows that lexical retrieval remains the primary path across all benchmarks, with lexical hit rates above $60\%$. 
This supports AdaPLD's lexical-first design, which preserves exact-match retrieval when available. 
Semantic fallback also provides substantial additional coverage, contributing more than $20\%$ absolute hit rate on every benchmark. 
Among lexical-miss cases, fallback recovers $67.37\%$--$80.07\%$, showing that it consistently reduces no-hit outcomes across both input-guided and reasoning benchmarks.

\subsection{Reuse Patterns Across Benchmarks}

\begin{table}[th]
\centering
\resizebox{0.5\textwidth}{!}{
\begin{tabular}{lccc}
\toprule
\bf Task / Benchmark& \bf Reuse Rate (\%)& \bf Branch-based Fraction (\%)& \bf Avg. Reuse Length\\
\midrule
\multicolumn{4}{l}{\textbf{Input-guided generation}} \\
 MT& 37.25& 44.92&3.65\\
 SUM& 60.33& 35.05&4.53\\
 RAG& 46.07& 47.19&3.59\\
 \midrule
 \multicolumn{4}{l}{\textbf{Input-guided editing}} \\
 Debug& 52.54& 40.53&4.20\\
 Polish& 42.57& 41.89&5.15\\
 Translate& 48.97& 41.21&3.21\\
 Switch& 58.00& 35.37&6.68\\
  \bottomrule
\end{tabular}
}
\caption{Reuse statistics aggregated by benchmark. Reuse Rate is the fraction of decoding steps with successful reuse. Branch-based Fraction is computed within reuse steps. Avg. Reuse Length denotes the mean accepted tokens per reuse step.}
\label{tab:ana_reuse_by_task}
\end{table}


Table~\ref{tab:ana_reuse_by_task} reports benchmark-level reuse statistics for input-guided generation and editing tasks. 
We summarize three aspects of reuse behavior: how often reuse succeeds, how often successful reuse involves branch-based paths, and how many tokens are accepted per successful reuse step.

\paragraph{Input-guided generation.}

For input-guided generation tasks, reuse rates vary from 37.25\% on MT to 60.33\% on SUM. 
This variation reflects differences in how strongly the generated output is grounded in the input context. 
Branch-based outcomes account for 35.05\%--47.19\% of successful reuse steps, showing that accepted reuse often goes beyond the direct main-path continuation. 
The average reuse length remains moderate, ranging from 3.59 to 4.53 tokens per reuse step.

\paragraph{Input-guided editing.}

Input-guided editing tasks show comparable reuse rates, ranging from 42.57\% to 58.00\%. 
The reuse length, however, varies more noticeably across editing tasks. 
Polish and Switch exhibit longer accepted spans, reaching 5.15 and 6.68 tokens respectively, while Translate has a shorter average reuse length of 3.21 tokens. 
This suggests that reuse is especially effective in editing tasks that preserve large portions of the input with local modifications, whereas stronger transformations provide shorter reusable spans.

\subsection{Accepted Reuse Length Analysis}
\label{app:accepted_reuse_length}

To further characterize reuse effectiveness, we analyze the distribution of accepted reuse length on CodeEditorBench with Vicuna-33B. For each successful reuse step, we define the accepted reuse length $\ell_{\mathrm{acc}}$ as the number of drafted tokens accepted by the target model before the first rejection.

\begin{table}[t]
\centering
\resizebox{0.5\textwidth}{!}{
\begin{tabular}{lccccc}
\toprule
Method & $\ell_{\mathrm{acc}}=1$ & $\ell_{\mathrm{acc}}=2$ & $\ell_{\mathrm{acc}}=3$ & $\ell_{\mathrm{acc}}=4$ & $\ell_{\mathrm{acc}}\geq 5$ \\
\midrule
PLD+   & 0.67 & 0.13 & 0.10 & 0.05 & 0.05 \\
AdaPLD & 0.46 & 0.31 & 0.13 & 0.05 & 0.05 \\
\bottomrule
\end{tabular}
}
\caption{Distribution of accepted reuse length on CodeEditorBench with Vicuna-33B.}
\label{tab:accepted_reuse_length}
\end{table}

Table~\ref{tab:accepted_reuse_length} shows the distribution of accepted draft-prefix lengths. 
Compared with PLD+, AdaPLD reduces one-token accepted prefixes from $0.67$ to $0.46$, while increasing prefixes with $\ell_{\mathrm{acc}}\ge 2$ from $0.33$ to $0.54$. 
Since $\ell_{\mathrm{acc}}=1$ advances decoding by only one token, this shift indicates that AdaPLD more often turns successful reuse into multi-token progress within a single verification step.

\subsection{Boundary Conditions for Effective Reuse}

\begin{figure}[th]
    \centering
    \includegraphics[width=\linewidth]{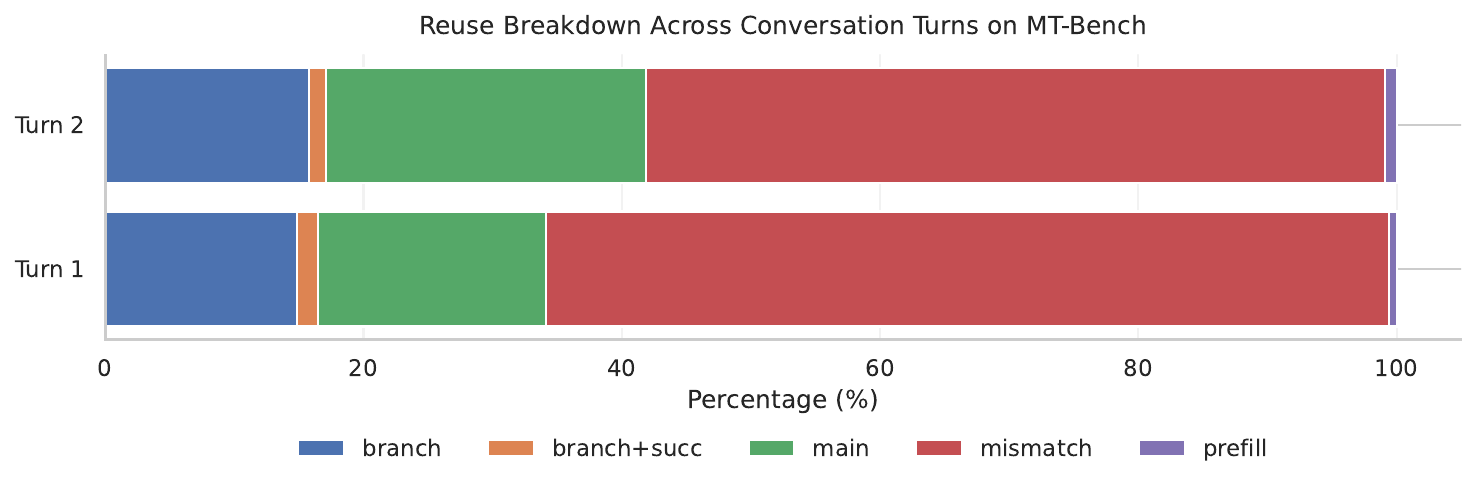}
    \caption{Comparison of reuse breakdown between Turn-1 and Turn-2 on MT-Bench, showing the proportion of different reuse outcomes in each turn.}

    \label{fig:ana_multi_turn}
\end{figure}

Reuse-based speculative decoding is most effective when the current generation is sufficiently supported by the available context, and when the additional draft search cost is outweighed by accepted-token gains. 
We analyze three factors that shape this trade-off: contextual grounding, draft search budget, and retrieval selectivity.

\paragraph{Contextual grounding.}

The availability of reusable context directly affects the success of speculative reuse. 
When generation is weakly grounded in the prompt or prior decoding history, retrieved anchors are less likely to support accepted continuations. 
We examine this effect using multi-turn conversation as a case study. 
Figure~\ref{fig:ana_multi_turn} compares reuse outcomes between the first and second turns on MT-Bench. 
In the first turn, the model has limited dialogue history to reuse, and speculative steps are more likely to produce no accepted draft tokens. 
In the second turn, the accumulated conversation history provides additional contextual structure, leading to more frequent reuse through both main-path and branch-based outcomes. 
This suggests that reuse becomes more effective as the available context becomes more informative for the current generation.

\paragraph{Draft search budget.}

The benefit of reuse also depends on how much draft structure is explored during verification. 
Increasing the branch width or maximum copy length can expose more reusable hypotheses, but it also increases verification cost. 
As shown in Figure~\ref{fig:abl_branch_width} and Figure~\ref{fig:abl_max_copy_length}, speedup improves only up to a moderate search budget and then decreases when the draft tree becomes too large or the copied span becomes too long. 
This indicates that effective reuse requires balancing exploration of additional continuations with the cost of verifying them.

\paragraph{Retrieval selectivity.}

Retrieval criteria determine how many candidate anchors are available for reuse. 
If the semantic similarity threshold is too restrictive, fallback retrieval may fail to recover useful candidates when lexical matching is sparse. 
As shown in Figure~\ref{fig:abl_emb_sim_threshold}, overly high thresholds reduce accepted reuse and can lower speedup. 
This supports the use of adaptive retrieval: lexical matching is preserved when available, while semantic fallback expands candidate coverage when strict token matching fails.

\begin{algorithm*}[t]
\caption{AdaPLD: Adaptive Retrieval and Reuse for Efficient Model-Free Speculative Decoding}
\label{alg:adapld}
\begin{algorithmic}[1]
\Require Target model $M$, maximum copy length $L$, branch width $K$, embedding threshold $\delta$, Context $\mathbf{x}_{\le i-1}$
\State $\mathbf{y}\gets[\,]$
\State \textbf{Prefill:} run $M$ on $\mathbf{x}_{\le i-1}$, append next token to $\mathbf{y}$ and update context
\While{not EOS}
    \State $x_{i-1}\gets$ \textit{bonus token} from the previous iteration
    \State $\mathcal{P}\gets \textsc{ExactMatch}(x_{i-1},\mathbf{x}_{\le i-1})$; \;\; $\textsf{strict}\gets (\lvert\mathcal{P}\rvert>0)$
    \If{$\lvert\mathcal{P}\rvert=0$}
        \State $\mathcal{P}\gets \textsc{SemanticFallback}(x_{i-1},\mathbf{x}_{\le i-1},\delta)$
        \If{$\lvert\mathcal{P}\rvert=0$}
            \State \textbf{AR step:} run $M$, append next token, update context; \textbf{continue}
        \EndIf
    \EndIf
    \State $p^\star \gets \textsc{RerankByHiddenState}(\mathcal{P},\mathbf{x}_{\le i-1},i)$

    \Comment{\textbf{Explicit draft construction}}
    \If{\textsf{strict}} \Comment{copy main path; branch excludes copied first token}
        \State $\mathbf{d}_{\text{main}} \gets \mathbf{x}_{p^\star+1:\min(p^\star+L,\,i-1)}$
        \State $\mathbf{d}_{\text{branch}} \gets \textsc{TopKNextTokens}(M,\mathbf{x}_{\le p^\star},K)\setminus\{\mathbf{d}_{\text{main}}[1]\}$
        \State $\mathbf{d}_{\text{succ}} \gets \varnothing$
    \Else \Comment{semantic hit: branch + successor}
        \State $\mathbf{d}_{\text{main}} \gets \varnothing$
        \State $\mathbf{d}_{\text{branch}} \gets \textsc{TopKNextTokens}(M,\mathbf{x}_{\le p^\star},K)$
        \State $\mathbf{d}_{\text{succ}} \gets \textsc{SuccessorDraft}(M,\mathbf{x}_{\le p^\star},\mathbf{d}_{\text{branch}},L)$
    \EndIf

    \State $\mathcal{T}\gets \textsc{BuildDraftTree}(\mathbf{d}_{\text{main}},\mathbf{d}_{\text{branch}},\mathbf{d}_{\text{succ}})$
    \State $\mathbf{z}\gets \textsc{VerifyAndAccept}(M,\mathbf{x}_{\le i-1},\mathcal{T})$
    \If{$\lvert\mathbf{z}\rvert=0$}
        \State \textbf{AR step:} run $M$, append next token, update context
    \Else
        \State append $\mathbf{z}$, update context (bonus token is $\text{last}(\mathbf{z})$)
    \EndIf
\EndWhile
\State \Return $\mathbf{y}$
\end{algorithmic}
\end{algorithm*}

\end{document}